\documentclass{article}
\usepackage{spconf,amsmath,epsfig, booktabs}

\title{Automatic Detection of Dark Ship-to-Ship Transfers using Deep Learning and Satellite Imagery}
\name{Ollie Ballinger}
\address{University College London\\	Centre for Advanced Spatial Analysis}

\begin{document}
%
\maketitle
\begin{abstract}
Despite extensive research into ship detection via remote sensing, no studies identify ship-to-ship transfers in satellite imagery. Given the importance of transshipment in illicit shipping practices, this is a significant gap. In what follows, I train a convolutional neural network to accurately detect 4 different types of cargo vessel and two different types of Ship-to-Ship transfer in PlanetScope satellite imagery. I then elaborate a pipeline for the automatic detection of suspected illicit ship-to-ship transfers by cross-referencing satellite detections with vessel borne GPS data. Finally, I apply this method to the Kerch Strait between Ukraine and Russia to identify over 400 dark transshipment events since 2022. 

\end{abstract}
\begin{keywords}
Deep Learning, YOLO, Ship Detection, Transshipment, Satellite Imagery 
\end{keywords}
\section{Introduction}
\label{sec:intro}

A Ship-to-Ship Transfer (STS) operation involves the exchange of cargo between two or more vessels while at sea. Though STS are often conducted for legitimate reasons, they are also one of the primary vehicles for concealing the origins of goods for the purposes of smuggling, sanctions evasion, and other illicit ends \cite{miller_identifying_2018}. “Dark” ship-to-ship transfers, in which one or more vessels involved turn off or manipulate their onboard Automatic Identification System (AIS) are of particular concern. Since the Russian invasion of Ukraine, sanctions evasion via dark transshipment has become so rife that the European Union banned access to EU ports for vessels which engage in ship-to-ship transfers if there is cause to suspect the cargo is of Russian origin \cite{european_comission_11th_2023}. 

There is a vast literature on the detection of ships using satellite imagery. The majority of these studies employ synthetic aperture radar imagery (SAR), which has a number of benefits for maritime surveillance including weather- and illumination-independent imaging \cite{wei_hrsid_2020, yasir_ship_2023, zhao_automatic_2022, zhao_domain_2023}. However, it is difficult to distinguish between different types of ship without optical information \cite{hong_multi-scale_2021}. Though a number of studies employ object detection or segmentation models to detect ships in optical satellite imagery, none of them identify transshipment \cite{corbane_complete_2010, kartal_ship_2019, li_ship_2021, mattyus_near_2013, proia_characterization_2010, yang_ship_2014, wu_inshore_2018}. 

Conversely, there have been several studies that automatically identify transshipment using vessel positions extracted from AIS data \cite{boerder_global_2018, masroeri_analysis_2022, miller_identifying_2018}. Ship-to-ship transfers are identified by algorithmically linking together vessels that loiter near each other for extended periods of time \cite{boerder_global_2018, miller_identifying_2018}. These studies have successfully shed light on illicit transshipment, mainly in the context of illegal fishing \cite{saito_transshipment_2022, kumar_identification_2022}. However, dark ship-to-ship transfers cannot be identified by relying solely on AIS data, as AIS is deliberately disabled to avoid detection \cite{bernabe_detecting_2023}. 

This study identifies dark transshipment using a three step procedure combining AIS data and satellite imagery. First, STS are identified in the AIS data by linking vessels that loiter together. The AIS data is then used to train an object detection model to identify STS in optical satellite imagery. Finally, STS detections in the satellite imagery are cross-referenced with AIS data to identify transshipment events that occur in the absence of AIS signatures. 

\begin{figure}[h!]
  \centering
  \includegraphics[width=\linewidth]{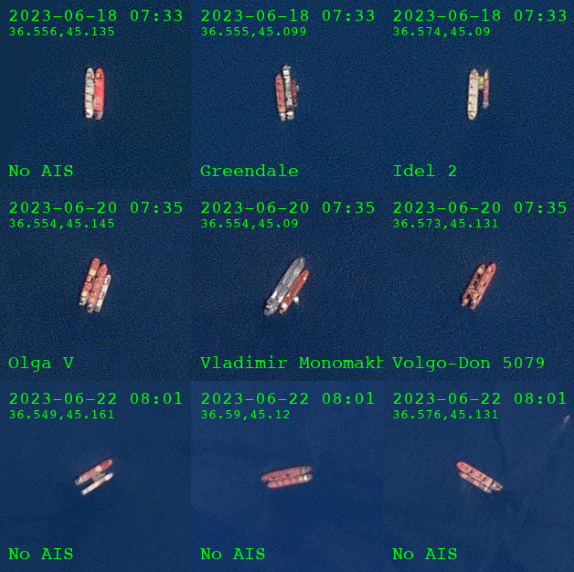}
  \caption{A sample of automatically detected dark STS}
\label{PS_AIS}
\end{figure}

\begin{figure*}[h!]  
  \centering
  \includegraphics[width=\textwidth]{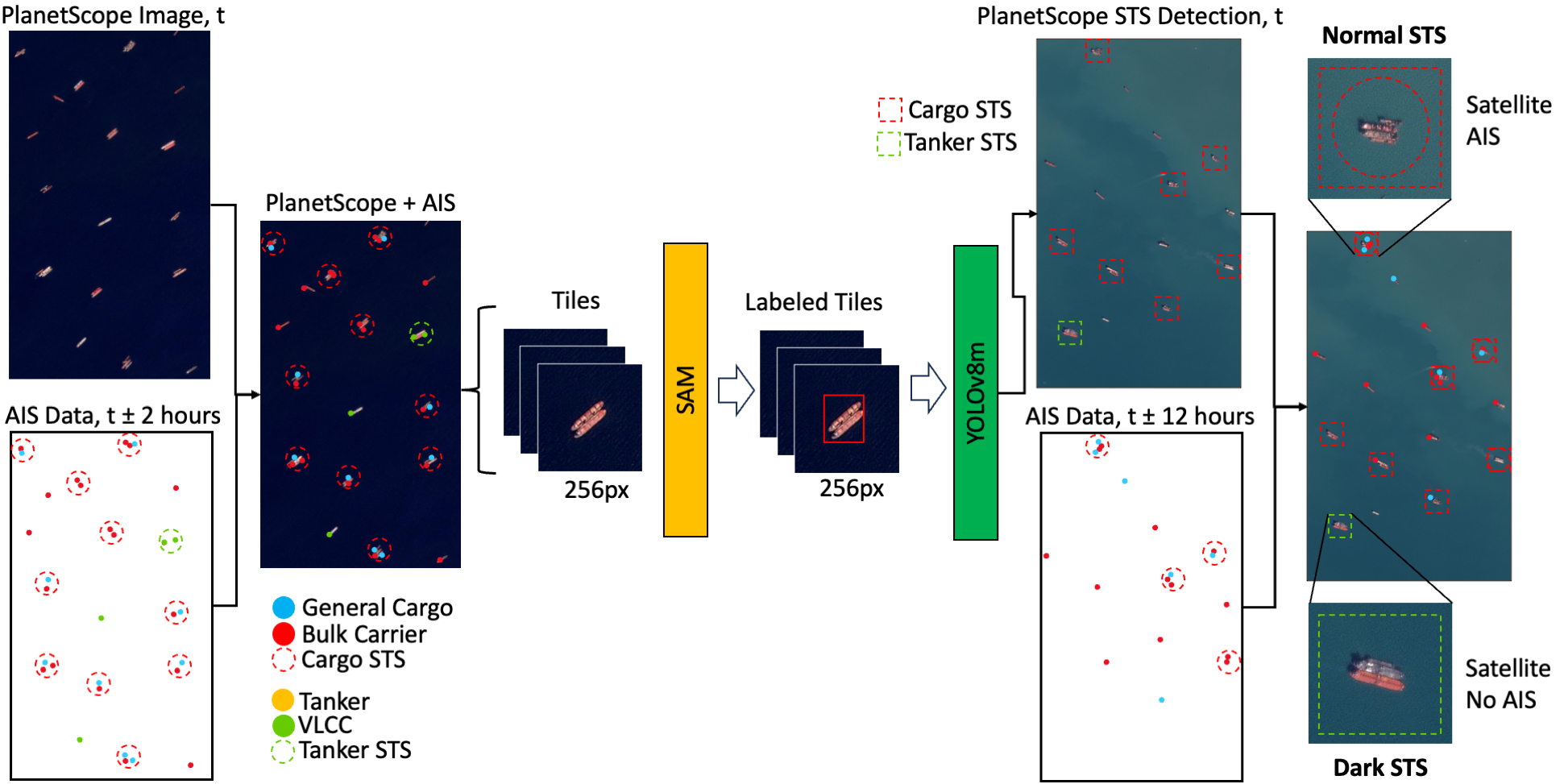}
  \caption{Dark Ship-to-Ship Transfer Detection Pipeline}
  \label{flowchart}
\end{figure*}

\section{Methodology}
\label{sec:methodology}

Figure \ref{flowchart} details the methodology employed herein. First, AIS data are used to spatiotemporally locate ships in satellite imagery. Coupled with zero-shot segmentation via the Segment Anything Model, this yields a labeled dataset which is used to train a YOLOv8m object detection model \cite{kirillov_segment_2023, redmon_you_2016}. Finally, object detection is carried out on new satellite imagery, and cross-referenced with AIS data. Ship-to-ship transfers detected in satellite imagery but where fewer than two AIS signatures are present are considered to be dark STS. 

\subsection{Data}
\label{ssec:Data}

Two main sources of data are employed in this study. The first is data collected from a ship’s Automatic Identification System (AIS), which encompasses vessel identification through unique International Maritime Organization numbers, real-time position data provided by GPS coordinates, and static details such as vessel dimensions and type. This information facilitates safe navigation, collision avoidance, and regulatory compliance in the maritime industry. AIS data were supplied by Lloyd’s List Intelligence, and included over 1 million unique vessel positions spanning between January 1st 2021 and September 1st 2023. 

The second data source is PlanetScope satellite imagery, which has a spatial resolution of 3 meters per pixel and a daily repeat cycle. Imagery was collected between January 1st, 2017 and September 1st, 2023. A cloud threshold of 0.7 was employed to discard cloudy scenes, resulting in imagery available for 531 days (55\% coverage). Both AIS data and satellite imagery were collected over the Kerch Strait anchorage (45\textdegree15'N, 36\textdegree30'E).
 
\subsection{Identifying STS in AIS data}

\label{ssec:Data}

AIS data are first used to locate ships spatiotemporally in the satellite imagery and label them into one of four classes;
\textbf{General Cargo:} A vessel capable of transporting a variety of goods with a capacity of up to 6,000 deadweight tons (DWT). \textbf{Bulk Carrier:} A ship designed to transport unpackaged dry bulk goods with a capacity exceeding 30,000 DWT. \textbf{Tanker:} A ship designed for the transportation of liquid cargoes with a capacity of up to 6,000 DWT. \textbf{VLCC (Very Large Crude Carrier):} A large tanker built for transporting crude oil with a capacity exceeding 100,000 DWT.

Then, transshipment is identified in AIS data by linking together ships that spend two or more hours within 500 metres of each other with a speed-over-ground below 1 knot. Depending on the type of ships involved in the transfer, the STS events are split into two categories; \textbf{STS Cargo} for transshipment involving dry cargo, and \textbf{STS Tanker} for transshipment involving liquid cargo (petroleum products).

This algorithm was applied to the full sample of AIS data, detecting over 12,000 ship-to-ship transfers in the Kerch strait since 2021, with more than half of these taking place since the Russian invasion of Ukraine.

\subsection{Cross-referencing AIS and Satellite Imagery}
\label{ssec:Data}

For each satellite image, the AIS location with the highest temporal proximity to the timing of the satellite image for each ship is taken, with a maximum time difference of 2 hours. A spatial buffer of 500 meters is then taken to generate an image tile containing a single ship or transshipment event. Figure \ref{PS_AIS} illustrates the procedure of cross-referencing AIS tracks and PlanetScope imagery. 

\begin{figure}[h!]
  \centering
  \includegraphics[width=\linewidth]{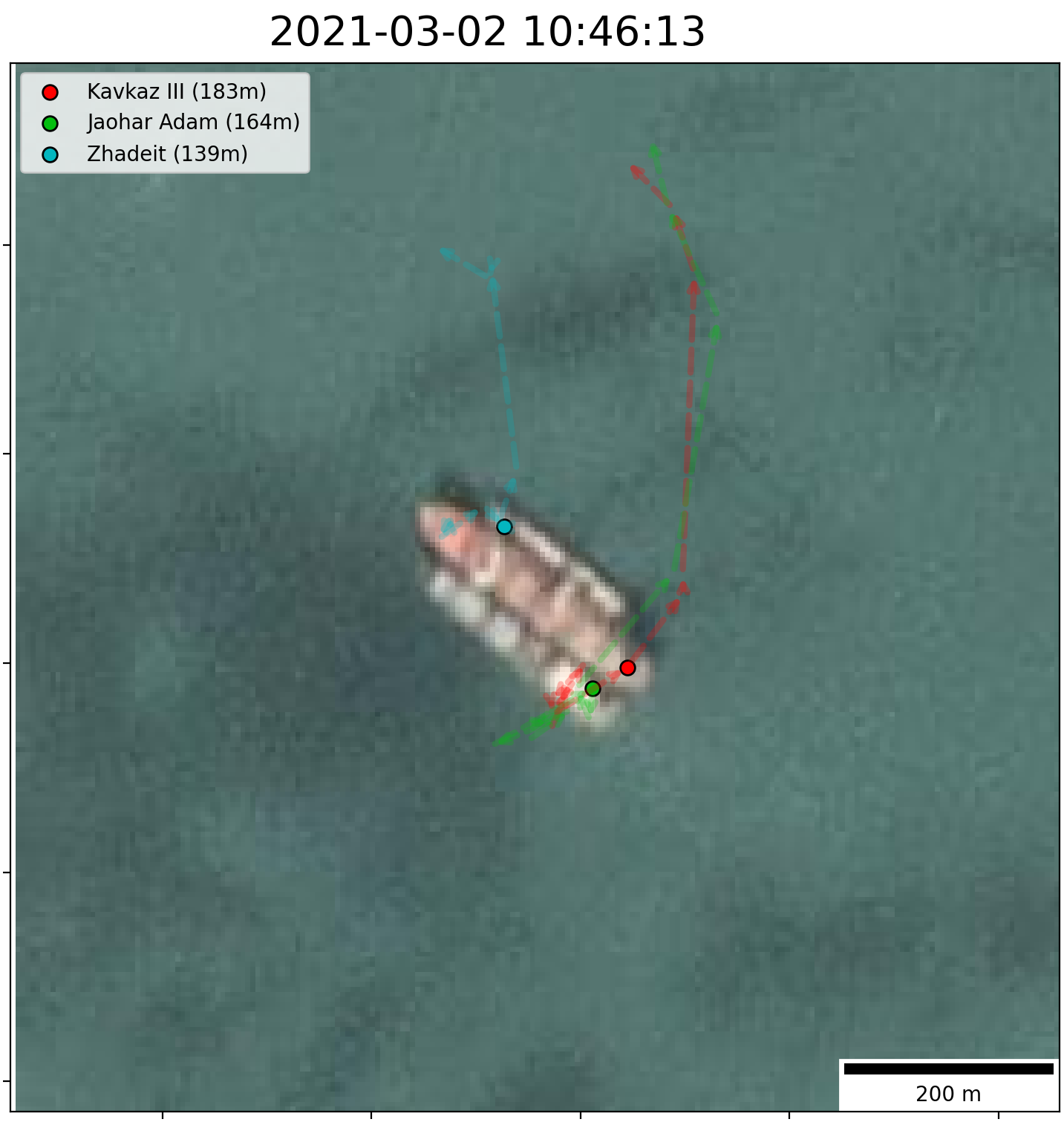}
  \caption{Cross-referencing AIS and Satellite Imagery}
\label{PS_AIS}
\end{figure}

Under normal circumstances, the agreement between the AIS data and the ships visible in satellite imagery is high enough that this procedure can not only localize the ship(s) in satellite imagery, but even the position of the AIS transponder on the bridge of the ship itself.

Once vessels have been located in the satellite imagery using the AIS data, the Segment Anything Model (SAM) developed by Kirilov et al. is employed to distinguish the ships themselves from the background \cite{kirillov_segment_2023}. High contrast between vessels and the background of open water enables efficient zero-shot segmentation. Segmentation masks are then converted to bounding boxes, and vessel information from the AIS data is used to label each image tile into one of six classes. Manual inspection of the result was conducted to ensure high quality. The class frequency distribution of the training dataset is reported in Table \ref{freq_table}.

\begin{table}
\centering
\begin{tabular}{lr}
\toprule
\textbf{Class} & \textbf{Frequency} \\
\midrule
General Cargo & 11995 \\
Bulk Carrier & 1946 \\
Cargo STS & 2081 \\
Tanker & 3303 \\
STS Tanker & 637 \\
VLCC & 261 \\
\midrule
\textbf{Total} & \textbf{20223} \\
\bottomrule
\end{tabular}
\caption{Frequency Distribution of Ship Classes}
\label{freq_table}
\end{table}

\subsection{Object Detection}
\label{ssec:Data}

This labeled dataset is then utilized to train a YOLOv8m object detection model using an 80-20 train-test split \cite{redmon_you_2016}. The model was trained for 50 epochs with a batch size of 128, and data augmentation techniques including mosaicking, random flips, and HSV color adjustments were employed. A Tesla P100 GPU with 16GB of memory was used for training, which lasted 3 hours and 17 minutes. 

\begin{figure}[h!]
  \centering
  \includegraphics[width=\linewidth]{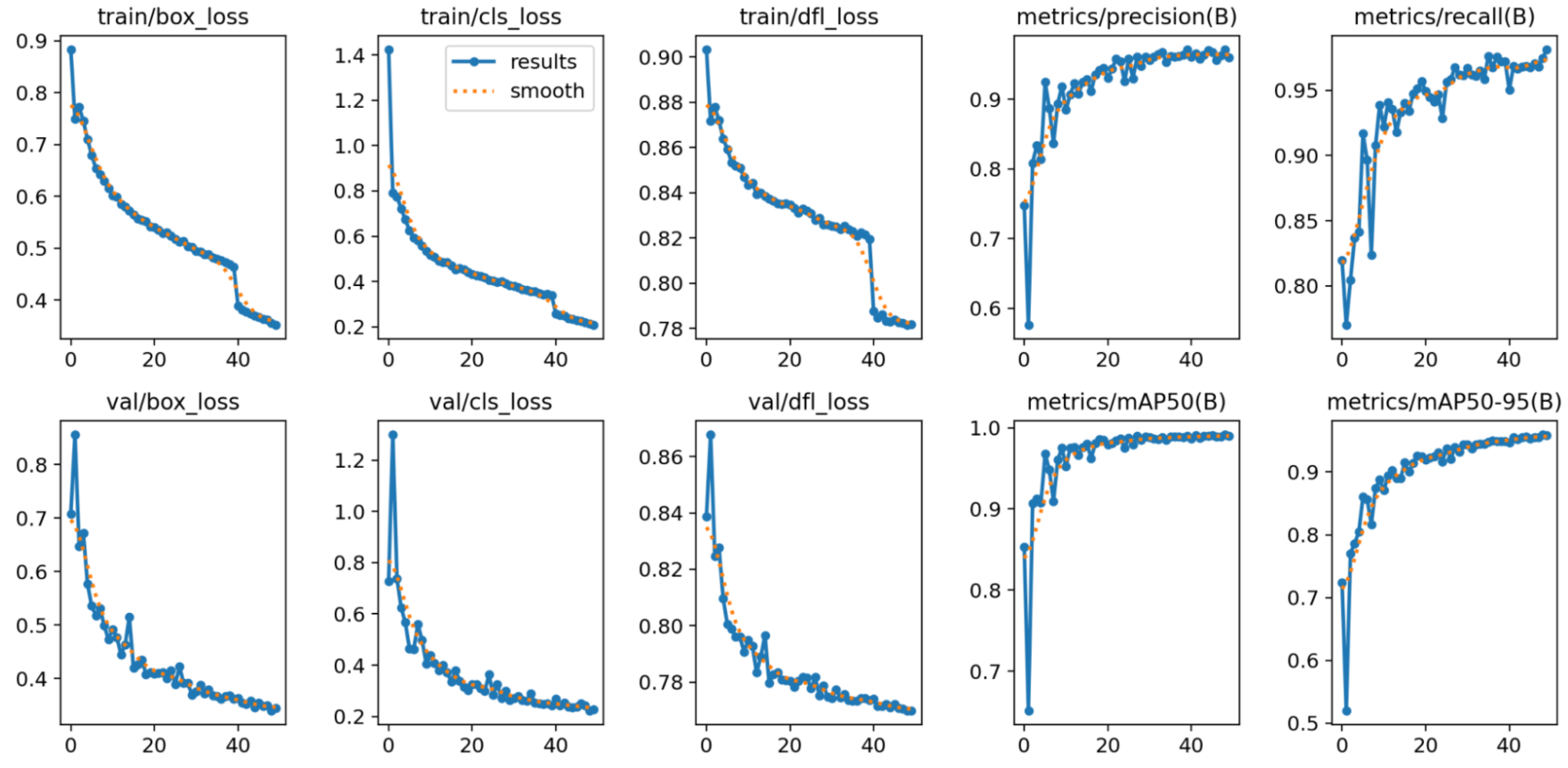}
  \caption{YOLOv8m model diagnostics}
\label{metrics}
\end{figure}

\begin{figure}[h!]
  \centering
  \includegraphics[width=\linewidth]{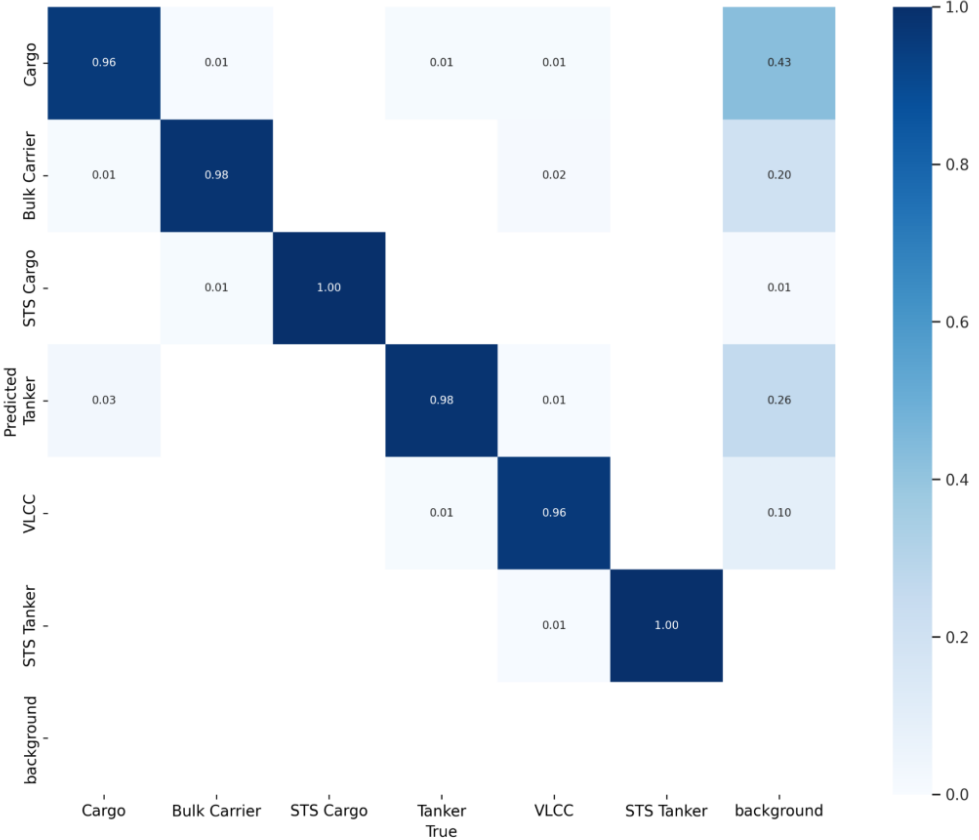}
  \caption{Normalized class-wise confusion matrix}
\label{cm}
\end{figure}

The model’s overall F1 accuracy is 97\%, and an mAP50 of 99\% indicating a high degree of accuracy in identifying the six classes of vessel in satellite imagery. Figure \ref{metrics} reports training diagnostics, and Figure \ref{cm} displays a normalized confusion matrix showing high class-wise accuracy in the test set. 

The trained model is not only highly accurate but light weight: at just 52MB (25.8m parameters) and with an inference speed of 2.179 ms per tile, it can search 115 km$^2$ per second. As such, it can be deployed without significant compute resources and analyze new imagery in near real time as it becomes available.

\section{Dark Ship-to-Ship Transfer Detection}
\label{sec:results}

\begin{figure}[h]
  \centering
  \includegraphics[width=\linewidth]{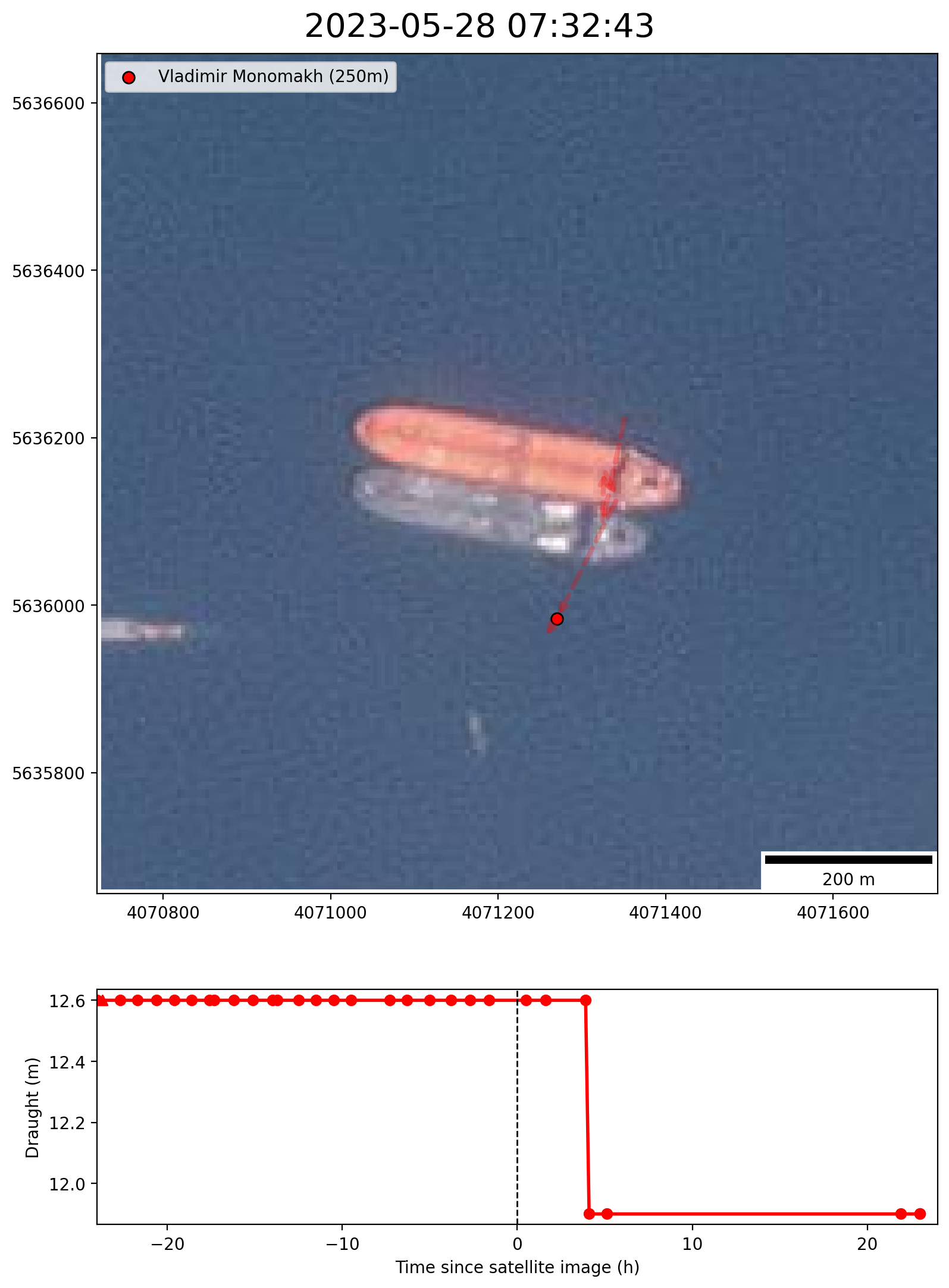}
  \caption{Dark STS involving Vladimir Monomakh}
\label{dark_vlad}
\end{figure}

To detect dark ship-to-ship transfers, the object detection model is first used to locate STS in satellite imagery. Once a transshipment event has been located, the surrounding 500m area is searched for AIS signals within 12 hours on either side of the satellite image. Because STS involves a minimum of two ships, searches which return less than two distinct vessel identities are classified as dark ship-to-ship transfers. In total, there were 402 dark ship to ship transfers identified in the Kerch Strait between 2021 and September 2023. Each one could involve the potential exchange of millions of dollars' worth of cargo. 

Figure \ref{dark_vlad} shows an example of an automatically detected dark STS. A transshipment event was detected in satellite imagery, but the corresponding location showed AIS transmissions from a single vessel in the 24 hours on either side of the satellite image. The one present AIS signal is from Vladimir Monomakh, a Very Large Crude Carrier owned by Russia's state oil company Rosneft with a length of 250 meters and a carrying capacity of 827,996 barrels \cite{british_petroleum_approximate_2021}. However, another even larger vessel is clearly visible in the satellite imagery. Vladimir Monomakh also experiences a draught change of 0.6 meters approximately two hours after this satellite image was taken, suggesting that the vessel lost significant weight.

Following Russia’s invasion of Ukraine, G7 countries instituted a price cap of \$60 per barrel for Russian crude oil, meaning that Vladimir Monomakh would be carrying roughly \$49 million worth of oil when at capacity. However, on the date of the transfer visible in Figure \ref{dark_vlad} the world price of brent crude was \$78 per barrel. As such, circumventing the price cap would increase the sale price of the ship’s cargo by around \$14 million. The world price of brent crude peaked at \$94/barrel in September 2023, increasing the value of this incentive to \$28 million.

Vladimir Monomakh was identified as having conducted 22 dark STS in which it had its own AIS turned on, but was detected in satellite imagery with another ship that had its AIS turned off. There are also a significant number of dark STS in which there are no AIS signals at all, but one of the ships present closely resembles Vladimir Monomakh. 

\section{Conclusion}
\label{sec:Conclusion}

There are a number of avenues for further research. The most obvious would be to expand the geographical coverage of the training data to improve out of sample performance. Another impactful extension of this research would be to conduct capacity estimation. This can be done straightforwardly, as there is a roughly log-linear relationship between ship length and carrying capacity, and the diagonal of a bounding box is a close approximation of a ship's length.

This study pioneers the application of deep learning techniques for the identification of dark Ship-to-Ship Transfers (STS) in satellite imagery, addressing a crucial gap in maritime surveillance. Leveraging a combination of Automatic Identification System (AIS) data and PlanetScope satellite imagery, the proposed methodology successfully detects dark STS events, where vessels manipulate or disable their AIS for illicit purposes. The results reveal over 400 dark transshipment events in the Kerch Strait between 2022 and September 2023, underscoring the prevalence of such deceptive maritime activities. The demonstrated efficacy of this approach signifies its potential for enhancing maritime security and facilitating timely intervention in illicit shipping practices.

\bibliographystyle{IEEEbib}
\bibliography{Ships_IEEE_refs}

\end{document}